\documentclass{article}
\usepackage{spconf,amsmath,graphicx,amsthm,amssymb,url}
\usepackage{cite} 
\usepackage{multirow}
\usepackage{algorithm}
\usepackage{algpseudocode}
\usepackage{colortbl}  
\usepackage{tabularx}
\usepackage[table,xcdraw]{xcolor}
\usepackage{booktabs}
\usepackage{subcaption}
\usepackage{CJKutf8}

\usepackage[utf8]{inputenc}
\usepackage[normalem]{ulem}

\title{Generative error correction for code-switching speech recognition using large language models}
\name{$^1$Chen Chen, $^1$Yuchen Hu, $^2$Chao-Han Huck Yang, $^1$Hexin Liu, \\ $^{2,3}$\emph{Sabato Marco Siniscalchi}, $^1$\emph{Eng Siong Chng}}
\address{$^1$Nanyang Technological University, Singapore; $^2$Georgia Institute of Technology, USA \\ $^3$Norwegian University of Science and Technology, Norway }
\begin{document}
\begin{CJK}{UTF8}{gbsn}

\maketitle
\begin{abstract}

Code-switching (CS) speech refers to the phenomenon of mixing two or more languages within the same sentence. Despite the recent advances in automatic speech recognition (ASR), CS-ASR is still a challenging task ought to the grammatical structure complexity of the phenomenon and the data scarcity of specific training corpus. In this work, we propose to leverage large language models (LLMs) and lists of hypotheses generated by an ASR to address the CS problem. Specifically, we first employ multiple well-trained ASR models for N-best hypotheses generation, with the aim of increasing the diverse and informative elements in the set of hypotheses. Next, we utilize the LLMs to learn the hypotheses-to-transcription (H2T) mapping by adding a trainable low-rank adapter. Such a generative error correction (GER) method directly predicts the accurate transcription according to its expert linguistic knowledge and N-best hypotheses, resulting in a paradigm shift from the traditional language model rescoring or error correction techniques. Experimental evidence demonstrates that GER significantly enhances CS-ASR accuracy, in terms of reduced mixed error rate (MER). Furthermore, LLMs show remarkable data efficiency for H2T learning, providing a potential solution to the data scarcity problem of CS-ASR in low-resource languages.   

\end{abstract}
\begin{figure*}[t]
\begin{center}
\includegraphics[scale=0.39]{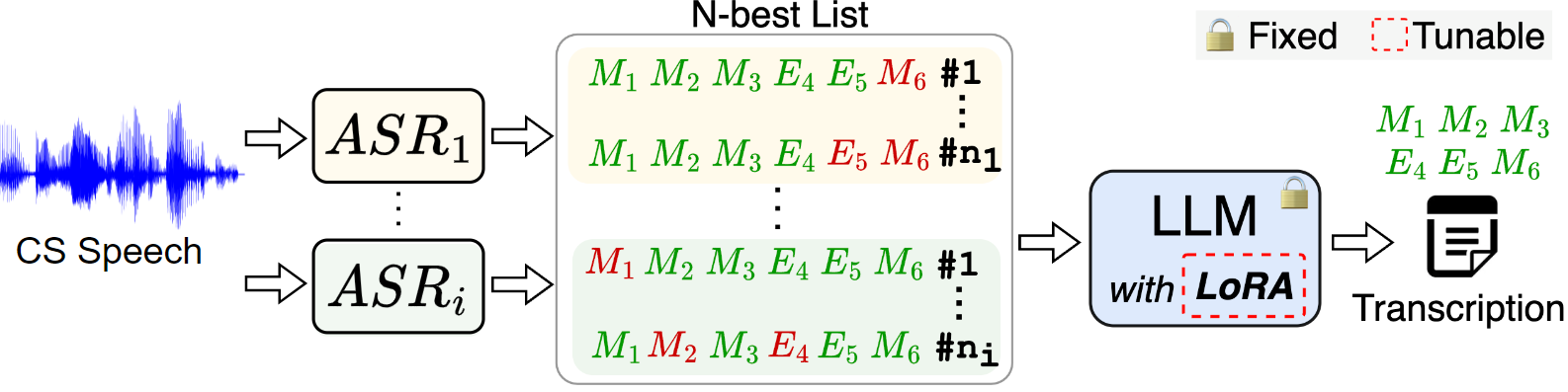}
\end{center}
\vspace{-0.2cm}
\caption{The main structure of the GER solution for Mandarin-English CS-ASR. $M_i$ and $E_i$ denote Mandarin and English tokens respectively. Correct tokens are highlighted in green, while wrong tokens appear in red. }
\vspace{-0.1cm}
\label{f1}
\end{figure*}
\section{Introduction}
\label{sec:intro}
The code-switching (CS) concept refers to the practice of alternating between more than one languages within a conversation or even within a single spoken utterance~\cite{yilmaz2016code}. This linguistic phenomenon is common in multilingual communities, and individuals who are fluent in multiple languages~\cite{nilep2006code}. CS poses a great challenge to automatic speech recognition (ASR) systems, which are typically trained on specific language resource~\cite{diwan2021multilingual, li2019towards,yue2019end, reduce_liu,modipa2013implications}. When CS occurs, ASR systems need to accurately transcribe the spoken item with multiple alternatives, since words from different languages may sound similar or have overlapping pronunciations~\cite{luo2018towards}. Considering the grammatical structure complexity, existing solutions focus on the technique of incorporating linguistic knowledge in the ASR engine~\cite{li2012code, adel2013recurrent, zhou2020end}, also known as internal LM estimation (ILME). More recently,~\cite{peng2022internal} has shown that an external LM can lead to a performance gain through rescoring the hypotheses generated by beam search. However, LM's potential for the CS task is still under-exploited. In fact, (i) rescoring-based methods select only one hypothesis as their output, which inevitably overlooks correct elements from other discarded hypotheses~\cite{wang2022effective}, and (ii) a less powerful language model may struggle to handle complex ambiguities and homophony when trained with limited in-domain data.\par
In this paper, we introduce a generative error correction (GER) approach that utilizes large LMs (LLMs) to directly predict the correct transcription for the CS-ASR task~\cite{chen2023hyporadise}. Different from traditional rescoring or correction techniques, GER can effectively exploit token-level information in the N-best hypotheses generated by an ASR system. Furthermore, LLMs have acquired vast knowledge of cross-lingual combination during the pre-training stage, so that the need for ad-hoc CS training data to build a robust ASR system is effectively reduced. Specifically, GER consists of two key steps: \par
\begin{itemize}
    \item Multiple existing ASR models are used to transcribe spoken utterances into an informative hypotheses list, thereby capturing more elements that potentially contribute to a more accurate transcription.
    \item A pre-trained foundation LLM with low-rank adapter\cite{hu2021lora} is employed to efficiently learn to predict transcription based on the hypotheses list, with limited in-domain hypotheses-transcription pairs.   
\end{itemize}
We conduct experiments on the English-Mandarin ASRU dataset\cite{shi2020asru}. The experimental results show that GER can significantly reduce the mixed error rate (MER) for each independent ASR system, and further enhance the performance after integrating their hypotheses lists. Furthermore, GER shows remarkable data efficiency when learning hypotheses-to-transcription (H2T) mapping. It requires less than 10\% in-domain data to build a better CS-ASR system than a Conformer-based baseline system. 

\section{Generative error correction}
In the following, we first define GER in Section~\ref{2-1}, and we then illustrate the generation of N-best hypotheses list in Section~\ref{2-2}. Finally, we introduce the low-rank adaptation (LoRA) method to learn the hypotheses-to-transcription (H2T) mapping.

\subsection{Problem Setting}\label{2-1}
\begin{figure}[t]
\begin{center}
\includegraphics[scale=0.9]{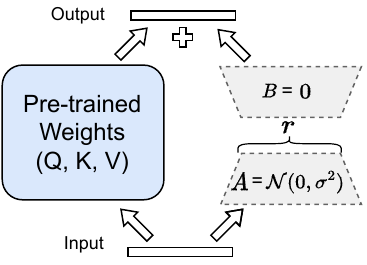}
\end{center}
\vspace{-0.1cm}
\caption{Low-rank adaptation. Only $\mathcal{A}$ and $\mathcal{B}$ are trainable. }
\vspace{-0.1cm}
\label{f2}
\end{figure}

As shown in Fig.~\ref{f1}, given input speech $X$ with CS, the $i$-th ASR model, $ASR_i$, transcribes $X$ to $n_i$ hypotheses list $\{Y_1^i,Y_2^i,\cdots,Y^i_{n_i}\}$, where each $Y$ can be viewed as an independent textual representation of $X$. After integrating candidates from different $ASR_i$, we can obtain an N-best list $\mathcal{Y}_N^I$ consisting of $N=\sum_{i=1}^I n_i$ hypotheses transcribed from $I$ ASR models. In GER, the goal is to learn a hypotheses-to-transcription mapping $\mathcal{M}_{H2T}$ that predicts the transcription $Y$ based on $\mathcal{Y}_N^I$ list, which can be expressed as:
\begin{equation}
    y = \mathcal{M}_{H2T} (\mathcal{Y}_N^I)
\end{equation}
Given ground-truth transcription $Y^*$, we can optimize the large language model to learn $\mathcal{M}_{H2T}$ in an auto-regressive manner. The cross-entropy loss $\mathcal{L}_{H2T}$ can be denoted as:
\begin{equation}
    \mathcal{L}_{H2T} = \sum_{t=1}^T -\log \mathcal{P}_{\theta} (y_t^* | y_{t-1}^*, \cdots, y_1^*, \mathcal{Y}_N^I)
\label{celoss}
\end{equation}
where $y_t$ is the $t$-th token of ground-truth transcription, and $\theta$ denotes the tunable module of LLM (i.e., LoRA in Fig.~\ref{f2}).

\subsection{Hypotheses Generation}~\label{2-2}
Mandarin-English is used as an example to illustrate the generation of an informative hypotheses list. To transcribe speech into textual candidates, we build a CS-ASR model ${ASR_1}$ from scratch and utilize two other existing models, i.e., ${ASR_2}$ (CS-ASR) and ${ASR_3}$ (monolingual). The underlying reason is that when applying ASR tasks in practical scenarios, we would simultaneously consider training a specific model with in-domain data, as well as seeking existing ASR models that can directly perform inference. It is noted that the hypothesis quality of ${ASR_1}$ relies on the in-domain data amount, while ${ASR_2}$ and ${ASR_3}$ are stable as they are well-trained. Beam search is employed in the decoding stage by all ASR models used to generate diverse hypotheses. The details for $ASR_i$ are as follows: \par

\noindent  $\boldsymbol {ASR_1}$-\textbf{Conformer} (48.3M) is a mainstream end-to-end ASR model with a remarkable performance~\cite{gulati2020conformer}. It is typically viewed as the primary solution to build a domain-specific ASR system. In practice, a Conformer consists of a 12-layer encoder and 6-layer decoder, and we utilize in-domain CS data to train it from scratch leveraging the ESPNet toolkit~\footnote{\url{https://github.com/espnet/espnet/blob/master/egs2}}. \par
\noindent$\boldsymbol {ASR_2}$-\textbf{Whisper}-Large (1.5B) is a web-scale speech model that is pre-trained on 680,000 hours of multilingual and multitask supervised data~\cite{radford2023robust}. We view it as a general ASR model since it has achieved remarkable ASR performance in various speech domains. To perform zero-shot CS speech recognition, we utilize the prompting technique in~\cite{peng2023prompting} to inspire Whisper as a hypotheses generator. \par  
\noindent$\boldsymbol {ASR_3}$-\textbf{Transformer} (31.0M) is a monolingual ASR model that is pre-trained on Mandarin corpus AISHELL-1~\cite{bu2017aishell}. This model consists of a 12-layer Transformer encoder and a 6-layer Transformer decoder that incorporates the CTC probabilities into decoding~\cite{dong2018speech}. We select such a monolingual ASR model to provide sufficiently potential Chinese tokens, as the target dataset is biased towards Mandarin. More importantly, the H2T learning based on ${ASR_3}$-Transformer is equivalent to \emph{transliteration}~\cite{yan2023towards}, where LLMs attempt to predict CS transcription according to Mandarin-only hypotheses list, e.g., ``他在等奥佛" $\rightarrow$``他在等offer". \par
We then combine the hypotheses generated from $ASR_1 \sim ASR_3$ to obtain a new N-best list that includes more useful elements for predicting the correct transcription. To validate our claim, we analyze the information on each $ASR_i$ as well as the ensemble hypotheses in our experiments, and we also conduct H2T learning on both single and ensemble N-best lists to evaluate the performance gain by LLMs.  

\subsection{H2T Learning with LoRA}~\label{2-3}
To learn the H2T mapping, we integrate a trainable low-rank decomposition matrix~\cite{hu2021lora} into LLMs’ existing Transformer layers (as shown in Fig.~\ref{f2}), enabling the model to adapt to new data while keeping the original LLMs fixed to retain the previous knowledge. Specifically, LoRA performs a reparameterization of each model layer expressed as a matrix multiplication by injecting low-rank decomposition matrices. As a result, the representations generated by the LLM are not distorted due to task-specific (H2T) tuning, while the adapter module acquires the capability to predict the true transcription from the N-best hypotheses~\cite{chen2023hyporadise}. \par
We use an instruction-following prompt template to formulate the N-best list as LLMs input, which is specified as: \emph{task description} + \emph{the best hypothesis} + \emph{other hypotheses}. Then LLMs perform auto-regressive decoding according to the input and history sequence, with the cross-entropy training criteria defined in Eq.(\ref{celoss}). 
 
\section{Experiments}
\subsection{Dataset}
We conduct our experiments on the ASRU dataset~\cite{shi2020asru}, which is a conversational Mandarin-English CS corpus collected by smartphones in quiet rooms. The transcripts of data cover many common fields, including entertainment, travel, daily life, and social interaction. The speakers are from 30 provinces in China, and 70 \% of them were under 30 years old, with no significant difference in the number of males and females. Each utterance contains 8.6 Chinese characters and 1.6 English words on average. \par
The original ASRU dataset consists of 200 hours of CS training set $Train_{CS}$ and 20 hours of evaluation set $Dev_{CS}$. There is no domain shift between $Train_{CS}$ and $Dev_{CS}$, so they are both defined as \textbf{\emph{in-domain data}}. In this paper, to show the efficiency of the proposed GER method, we re-split the $Dev_{CS}$ to simulate the low-resource setting, where 15k utterances (19 hours) are randomly selected as low-resource training set and other unseen 1152 samples (1.5 hours) are viewed as a new test set to evacuate the performance. 

\subsection{LLM-LoRA Configuration}
We use Chinese-Llama2-7b from HuggingFace~\footnote{https://huggingface.co/ziqingyang/chinese-llama-2-7b} as the foundation model to learn H2T mapping, which has been expanded and optimized with Chinese vocabulary beyond the original Llama-2~\footnote{https://ai.meta.com/llama/}. Consequently, the subsequent training on Chinese corpus improves the fundamental semantic understanding of both Chinese and English, which is suitable for CS H2T learning. For LoRA training, $r$ is set to 4 and added in each Q, K, and V layers in the Transformer blocks. Notably, the trainable parameters are only 4.19M, accounting for only 0.06\% of the Llama parameters. The learning rate is set to $2\times e^{-4}$ to learn H2T mapping for 10 epochs, and the batch size is set to 128. 

\subsection{Evaluation Indicators}\label{3-3}
We utilize the mix error rate (MER) as the metric to evaluate the performance of each ASR system. To gain more insights into our results, we also define two oracles (\romannumeral1) $o_{nb}$, and (\romannumeral2) $o_{cp}$ for a given N-best list. $o_{nb}$ denotes the WER of the best hypothesis in the N-best list. We measure $o_{nb}$ by calculating MER of all hypotheses one by one. $o_{cp}$ denotes the theoretical minimum WER that recomposes all tokens in the N-best list. We measure $ \boldsymbol o_{cp}$ by counting the rate of missing tokens in N-best list. Additionally, we use ``1-\emph{Best}" to denote the WER of the best hypotheses in the list.
\begin{table*}[t]
\centering
\begin{tabular}{c|c|cc|c|c|c|cc}
\toprule[1.5pt]
\multirow{2}{*}{ID}& \multirow{2}{*}{ASR Model} & \multicolumn{2}{c|}{Data Dur. (h)} & \multirow{2}{*}{Language} & \multirow{2}{*}{1-\emph{Best}} & \multirow{2}{*}{\textbf{GER}} & \multicolumn{2}{c}{\emph{Oracle}} \\
& & In-D. & Out-D.&  & & & $o_{nb}$ & $o_{cp}$ \\ \midrule
1&\multirow{2}{*}{$ASR_1$-Conformer} & 19 & - & \multirow{2}{*}{M\&E} &37.9 & - & 33.3 & 29.9 \\
2&& 200 & - & &12.9 & $10.6_{\textcolor{teal}{-17.8\%}}$\ & 8.9 & 7.6\\ \midrule
3&$ASR_2$-Whisper &19 & 680k & Multi. & 11.0 &$8.6_{\textcolor{teal}{-20.9\%}}$ & 9.5 & 7.8 \\ \midrule
4&$ASR_3$-Transformer &19 & 150 & M & 41.1 & ${30.8}_{\textcolor{teal}{-25.1\%}}$ &38.1 &22.8 \\ \midrule
5&Ensemble & 19 & - &  M\&E & 11.0 & $\textbf{8.3}_{\textcolor{teal}{-24.5\%}}$ & 8.6& 6.5 \\
\bottomrule[1.5pt]
\end{tabular}
\caption{Main MER (\%) results on ASRU test set. ``In-D." and ``Out-D." respectively denote in-domain and out-domain training data, and ``M" and ``E" stands for Mandarin and English. ``1-\emph{Best}", ``$o_{nb}$" and ``$o_{cp}$" are all defined in Section~\ref{3-3}.}
\vspace{-0.1cm}
\label{t1}
\end{table*}
\section{Results and Analysis}
\subsection{Main Results of GER}
Table~\ref{t1} shows the main MER results. For each $ASR_i$, we select the top-5 hypotheses to compose the N-best list for both oracle analysis and H2T learning. Considering the MER result of each model, the ensemble N-best list includes 8 hypotheses from ID 1,3,4 with a proportion of 1:5:1. Additionally, it is worth noting that only ID-2 use full training set $Train_{CS}$, all other systems except utilize 19 hours of in-domain training data for the H2T learning, which accounts for \textbf{less than 10\%} of other typical methods trained on ASRU. \par
From Table~\ref{t1}, we observe that: 1) The ASR model with ID-1 shows that 19 hours of training data is not enough to train the $ASR_1$ from scratch, therefore, it cannot provide high-quality hypotheses for GER learning. 2) With 200 hours of training data (ID-2), the Conformer baseline achieves 12.9 MER performance while the GER can further reduce the relative MER by 17.8\%. 3) Whisper can achieve better performance than Conformer due to its large-scale pre-trained data. Nonetheless, GER reaps significant performance gain (11.0\% $\rightarrow$ 8.6) using 19 hours of in-domain data. 4) $ASR_3$ achieves poor MER performance as it is a monolingual ASR model, and GER can perform transliteration with reducing MER by 25.1\%, which predicts Mandarin-English transcription according to Mandarin-only hypotheses. 5) When integrating the hypotheses into an ensemble list, $o_{np}$ and $o_{cp}$ both drop, which indicates a higher upper bound for GER. Accordingly, the GER obtains better results than the single model that surpasses the 1-\emph{best} baseline by 24.5\% MER performance. In general, GER can enhance MER for both stand-alone and ensemble ASR models. Furthermore, GER with Whisper and 19 hours of training data allows us to build a better CS-ASR system than a Conformer baseline trained on 200 hours.

\begin{table}[t]
\centering
\begin{tabular}{c|cc|c|cc}
\toprule[1.5pt]
\multirow{2}{*}{1-\emph{Best}} &\multicolumn{2}{c|}{Data Amount}  & \multirow{2}{*}{GER} & \multicolumn{2}{c}{Oracle}   \\ 
& \#Pairs & Dur. & & $o_{nb}$& $o_{cp}$\\\midrule
\multirow{4}{*}{11.0}& 1k & 1.3 h & $10.1_{\textcolor{teal}{-8.2\%}}$ & \multirow{4}{*}{9.4} & \multirow{4}{*}{7.0} \\
&5k & 6.3 h & $9.3_{\textcolor{teal}{-15.5\%}}$ & &\\
&10k & 12.6 h & $8.9_{\textcolor{teal}{-19.1\%}}$ & &\\
&15k & 19.0 h  &$8.6_{\textcolor{teal}{-20.9\%}}$& & \\
 
\bottomrule[1.5pt]
\end{tabular}
\caption{MER (\%) results of GER method with different amounts of in-domain training data. ``\#Pairs" denotes the number of hypotheses-transcription pairs, and ``Dur." represents the corresponding speech duration. }
\vspace{-0.15cm}
\label{t2}
\end{table}

\subsection{Data Efficiency}
We now examine the data efficiency of GER by gradually reducing the amount of in-domain training data. As shown in the second column of Table~\ref{t2}, we generate hypotheses-transcription pairs using $ASR_2$ and $ASR_3$ with a proportion of 5:1. We do not use $ASR_1$ since it cannot provide reasonable hypotheses with such a small amount of training data. \par
From Table~\ref{t2}, 
we observe that the relative MER can be reduced by 8.2\% by using only 1K in-domain pairs (1.3 hours) for H2T learning.
This outcome demonstrates that a multi-lingual LLM can significantly reduce the data requirement of in-domain data for CS-ASR training. We believe it provides a potential solution for those languages with limited CS training corpus.

\subsection{Case Study}
We conduct the case study in Table~\ref{t3} to visualize the error correction process. All three ASR models provide defective hypotheses with different errors. Nevertheless, as all correct tokens are included in N-best list, GER with LLM accurately predicts the true transcription according to semantics and grammar of Mandarin and English: 1) ``persistent data" is the most likely object to be invented in this context, and 2) the subject of the invention should be ``他" rather than ``它" according to Chinese grammar. 
\begin{table}[t]
\centering
\resizebox{1.0\columnwidth}{!}{\begin{tabular}{c|c|c}
\toprule[1.5pt]
System & Utterance & MER \\ \midrule
$ASR_1$& persistent \textcolor{red}{date}这个东西当然不是他发明的 &7.1 \\ \midrule
$ASR_2$& \textcolor{red}{porsistent} data这个东西当然不是\textcolor{red}{它}发明的  &14.3\\  \midrule
$ASR_3$&\textcolor{red}{颇虽私人的队的}这个东西当然不是他发明的  &50.0 \\ \midrule
GER &\textcolor{teal}{persistent data}这个东西当然不是\textcolor{teal}{他}发明的& 0 \\ \midrule
GT &persistent data这个东西当然不是他发明的& -\\
\bottomrule[1.5pt]
\end{tabular}}
\caption{Case study of GER. The utterance is drawn from the ASRU test set, where error tokens are in red. }
\vspace{-0.15cm}
\label{t3}
\end{table}

\section{Conclusion}
In this paper, we present a new paradigm called generative error correction (GER) for CS-ASR. This GER approach first employs well-trained ASR models to generate diverse hypotheses, and then learns an H2T mapping using LLMs with LoRA adapter. The experiments show that GER can effectively enhance the CS-ASR in terms of MER. Meanwhile, it shows remarkable data efficiency that provides a potential solution to the data scarcity problem of CS-ASR in low-resource language.  
\newpage

\bibliographystyle{IEEEtran}
\bibliography{refs}
\end{CJK}
\end{document}